\documentclass[10pt,twocolumn,letterpaper]{article}

\usepackage[pagenumbers]{cvpr}

\usepackage{algorithm}
\usepackage{algorithmic}
\usepackage{colortbl}
\usepackage{multirow}
\usepackage{tabularx}
\usepackage{listings}
\usepackage{newfloat}
\usepackage[most]{tcolorbox}
\usepackage{titlesec}
\usepackage{cuted}
\usepackage{balance}
\usepackage{placeins}
\usepackage{float}

\definecolor{TechNavy}{HTML}{143A5A}
\definecolor{TechBlue}{HTML}{2E75B6}
\definecolor{TechSky}{HTML}{74B9E6}
\definecolor{TechMist}{HTML}{EAF6FD}
\definecolor{TechRule}{HTML}{B9DDF2}
\usepackage[breaklinks,colorlinks,linkcolor=TechBlue,citecolor=TechBlue,urlcolor=TechBlue]{hyperref}

\titleformat{\section}
  {\large\bfseries\color{TechNavy}}
  {\thesection}{0.55em}{}
  [\vspace{0.12em}{\color{TechRule}\titlerule}]
\titleformat{\subsection}
  {\normalsize\bfseries\color{TechNavy}}
  {\thesubsection}{0.5em}{}
\titleformat{\subsubsection}
  {\normalsize\bfseries\color{TechBlue}}
  {\thesubsubsection}{0.5em}{}

\newcommand{\method}{\textsc{Branch2Skill}}

\newcolumntype{Y}{>{\raggedright\arraybackslash}X}
\newcommand{\equalcontrib}{\textsuperscript{*}}
\newcommand{\corresponding}{\textsuperscript{\textdagger}}
\newcommand{\reportaffiliations}{}
\newcommand{\affiliations}[1]{\gdef\reportaffiliations{#1}}
\makeatletter
\let\report@maketitle\maketitle
\renewcommand{\maketitle}{%
  \g@addto@macro\@author{\\[0.7em]\reportaffiliations}%
  \report@maketitle
}
\makeatother

\lstdefinestyle{promptstyle}{
  basicstyle=\footnotesize\ttfamily,
  breaklines=true,
  breakatwhitespace=false,
  columns=fullflexible,
  keepspaces=true,
  showstringspaces=false,
  tabsize=2,
  aboveskip=0pt,
  belowskip=0pt
}
\DeclareCaptionStyle{ruled}{labelfont=normalfont,labelsep=colon,strut=off}
\DeclareFloatingEnvironment[fileext=lst,placement=tb,name=Listing]{listing}
\allowdisplaybreaks
\title{\color{TechNavy}Branch2Skill: Efficient Skill Evolution Through Reasoning Trees}
\author{
    Yanwei Ren\textsuperscript{1,2}\quad
    Haotian Zhang\textsuperscript{1,2}\quad
    Likang Xiao\textsuperscript{2}\quad
    Jiaxing Huang\textsuperscript{3}\\
    Jiayan Qiu\textsuperscript{4}\quad
    Baosheng Yu\textsuperscript{5}\quad
    Quan Chen\textsuperscript{6}\quad
    Liu Liu\textsuperscript{1,2}\corresponding
}
\affiliations{
    \textsuperscript{1}School of Artificial Intelligence, Beihang University, Beijing, China\\
    \textsuperscript{2}Hangzhou International Innovation Institute, Beihang University, Hangzhou, China\\
    \textsuperscript{3}The Hong Kong Polytechnic University, Hong Kong SAR, China\\
    \textsuperscript{4}University of Leicester, Leicester, United Kingdom\\
    \textsuperscript{5}Nanyang Technological University, Singapore\quad
    \textsuperscript{6}Kuaishou Technology, Beijing, China\\
    \corresponding\ Corresponding author: liuliubh@buaa.edu.cn
}

\begin{document}

\raggedbottom

\maketitle

\begin{strip}
\centering
\begin{tcolorbox}[
  enhanced,
  width=0.94\textwidth,
  colback=TechMist,
  colframe=TechRule,
  boxrule=0.55pt,
  arc=2.2mm,
  left=1.4mm,
  right=1.4mm,
  top=1.2mm,
  bottom=1.2mm,
  drop fuzzy shadow=black!12,
  before skip=0.6em,
  after skip=1.0em
]
\begin{abstract}
Skill evolution improves agent skills through feedback over time, with failed trajectories often providing informative signals by revealing incomplete or misleading behaviors. However, existing methods mainly rely on single trajectories, where early reasoning errors can propagate through subsequent steps and weaken the feedback available for skill refinement. Consequently, improving skills requires repeated cycles of rollout, diagnosis, and update, incurring substantial token costs. To address this challenge, we introduce \textsc{Branch2Skill}, an efficient framework that transforms a single reasoning tree into dense supervision for skill evolution. For each task or problem, Branch2Skill performs Monte Carlo tree search under a fixed budget to obtain diverse reasoning trajectories, then compares an elite path with sibling alternatives sharing the same prefixes to extract step-wise evidence about which reasoning patterns to retain, revise, or avoid. Finally, Branch2Skill distills multi-step evidence into reusable updates, allowing one reasoning tree to provide supervision across multiple reasoning steps and reducing the need for repeated rollout-update cycles. 
Across six benchmarks covering reasoning and agentic tasks, \textsc{Branch2Skill} consistently improves task performance while enhancing skill evolution efficiency.
For example, with GPT 5.5 as the target model, \textsc{Branch2Skill} uses $73.2\%$ fewer tokens than SkillOpt,  while achieving superior performance. These results demonstrate that reasoning trees can support not only more effective trajectory search, but also richer supervision for more efficient skill improvement.
Code will be published.
\end{abstract}
\end{tcolorbox}
\end{strip}

\renewcommand{\method}{\textsc{Branch2Skill}}
\section{Introduction}
\label{sec:introduction}

LLM agents are increasingly expected to solve diverse tasks by accumulating experience and improving their capabilities over time. However, updating model parameters after every interaction is costly and impractical, motivating approaches that externalize learning through reusable artifacts. Verbal reflection and experiential memory allow frozen models to reuse useful procedures across tasks \citep{shinn2023reflexion,zhao2024expel}, while recent work further treats skills as objects that can be revised based on execution feedback \citep{yang2026skillopt,yu2026skilladaptor,chen2026skillcat}. Execution trajectories provide valuable signals: successful trajectories demonstrate that a given reasoning strategy can produce valid outcomes, while failed trajectories can be more informative for skill evolution by revealing where the strategy is incomplete, misleading, or insufficient. However, extracting actionable lessons from trajectories remains challenging. 






\begin{figure}[t]
    \centering
    \includegraphics[width=\columnwidth]{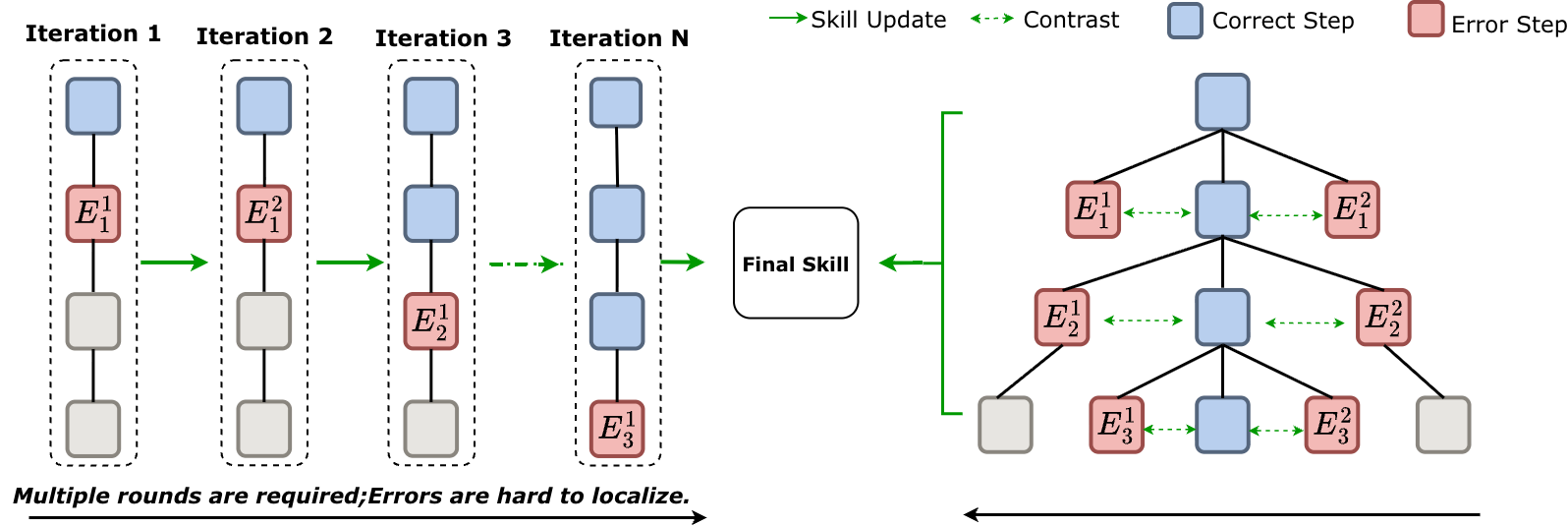}
    \caption{\textbf{Reasoning trees provide dense evidence.} Left, a failed rollout often yields only one reliable update signal
    because later steps inherit the first consequential error. Extracting
    another signal therefore requires a new generation and update round.
    Right, \textsc{Branch2Skill} compares each selected step with sibling
    alternatives sharing the same prefix across multiple depths. These local
    comparisons allow a single search tree to produce multiple skill-update signals.}
    \label{fig:motivation}
\end{figure}

Current methods commonly extract evidence from single trajectories \citep{yang2026skillopt,chen2026skillcat}. Such updates improve skills, but most generated reasoning does not provide independent update signals. A single rollout exposes only one path through the model's reasoning distribution. Once an early decision leads to a flawed reasoning state, later steps may simply inherit the original false premise or reflect local repairs rather than actionable lessons. Treating these steps as separate feedback risks encoding symptoms instead of addressing the source. Although locating the first actionable fault reduces ambiguity, the evidence remains limited to a single trajectory \citep{yu2026skilladaptor}. Obtaining additional failure signals requires new samples, diagnoses, or update cycles, resulting in sparse supervision and high token costs. Figure~\ref{fig:motivation} illustrates this limitation.


\begin{figure}[t]
    \centering
    \includegraphics[width=1\columnwidth]{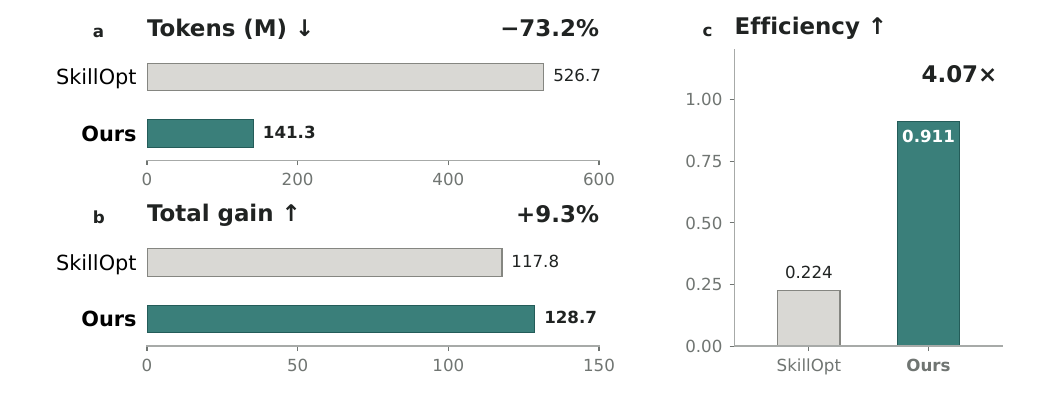}
    \caption{Aggregate performance and skill-evolution efficiency
    across the evaluated settings.} 
    \label{fig:efficiency}
\end{figure}

Monte Carlo tree search (MCTS) provides a richer feedback source by exploring alternative continuations and repeatedly expanding promising prefixes \citep{kocsis2006bandit}. When applied to reasoning tasks, tree search avoids committing to the first sampled path and reveals multiple possible decisions under the same partial solution~\citep{yao2023tree,zhou2024lats}. Compared with single rollouts, it thus captures a broader range of reasoning behaviors for the given task.
This advantage arises because sibling nodes share the same prefix and diverge only at the current decision, allowing their downstream outcomes to provide localized evidence about how alternative choices affect subsequent reasoning. Such comparisons are naturally induced by the tree
structure and search outcomes, without requiring an additional error-attribution model. Prior work has demonstrated the value of
intermediate-step supervision \citep{lightman2024verify}, while later work shows that sibling branches can refine reasoning trajectories~\citep{ren2025sigma}. However, how evidence distributed across a reasoning tree can be transformed into updates for a persistent external skill remains unexplored.

In this paper, we introduce \textsc{Branch2Skill}, a efficient search-guided framework that converts a reasoning tree into dense supervision for skill evolution. Given a training task and the current skills, \textsc{Branch2Skill} runs MCTS and selects an elite solution path. At each expanded decision point along this path, it compares the selected reasoning step with sibling alternatives that share the same prefix, using their downstream outcomes to determine which reasoning behaviors should be retained, revised, or avoided. The method consolidates evidence from different depths and distills it
into reusable updates. A single reasoning tree can therefore expose
multiple failure modes under matched reasoning contexts, rather than producing a single diagnosis from an individual rollout. Search branches created to solve the current problem become evidence for improving future behavior.

Across reasoning and agentic tasks, \textsc{Branch2Skill} improves both task
performance and skill-evolution efficiency. On LiveMathematicianBench
, it achieves a higher score than SkillOpt
\citep{yang2026skillopt} while using fewer skill-evolution tokens. Across the six benchmarks with GPT-5.5 as the target model,
\textsc{Branch2Skill} uses $73.2\%$ fewer skill-evolution tokens
than our SkillOpt reproduction while increasing the aggregate
test gain from $117.8$ to $128.7$ percentage points. Figure~\ref{fig:efficiency}
shows that token use falls from $526.7$ million to $141.3$ million, while total
gain rises from $117.8$ to $128.7$. This corresponds to $73.2\%$ fewer tokens,
$9.3\%$ greater total gain, and a $4.07$-fold increase in gain per million
tokens. These results support a broader role for reasoning search.


\section{Related Works}
\label{sec:related_work}

\begin{figure*}[!t]
    \centering
    \includegraphics[width=0.98\textwidth]{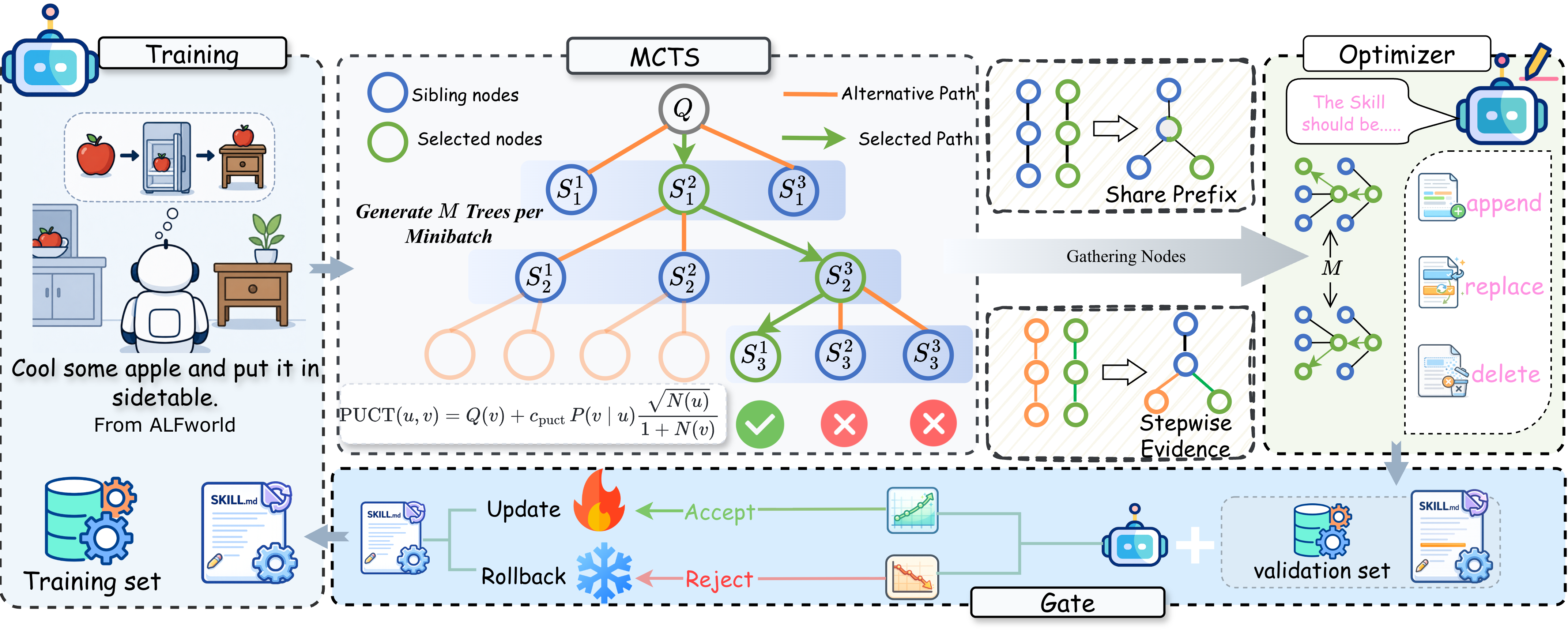}
    \caption{\textbf{Overview of \textsc{Branch2Skill}.}
    At each update step, the current skill is held fixed while the target
    model constructs one reasoning tree for each sampled problem. PUCT
    allocates search between high scoring nodes and less visited alternatives.
    After search, the method collects an elite path and sibling nodes that
    share a parent with each selected step. The skill model (optimizer)
    revises the current skill with \textsc{Append}, \textsc{Replace}, or
    \textsc{Delete} operations. Direct evaluation on a fixed validation set
    accepts a candidate when performance does not decline and otherwise
    restores the previous skill.}
    \label{fig:method_overview}
\end{figure*}

\paragraph{External experience.}
Language model agents can improve across tasks by storing experience outside
their parameters. Reflexion records verbal feedback in episodic memory, whereas
ExpeL extracts general lessons from agent trajectories
\cite{shinn2023reflexion,zhao2024expel}. Voyager maintains an executable skill
library for embodied interaction, and Buffer of Thoughts retrieves and updates
reusable reasoning templates \cite{wang2023voyager,yang2024buffer}. These
studies show that natural-language rules, programs, and reasoning patterns can
transfer experience without model training, but emphasize artifact construction
and reuse. \textsc{Branch2Skill} instead examines the evidence for revising a
persistent skill.

\paragraph{Skill synthesis and evolution.}
Recent work optimizes skills from completed trajectories. Trace2Skill
consolidates local lessons, whereas EvoSkill iteratively proposes and selects
edits \cite{ni2026trace2skill,alzubi2026evoskill}. SkillOpt applies bounded
edits with held-out validation, while SkillOpt-Lite combines trajectory
exploration, consensus mining, and validation gating
\cite{yang2026skillopt,shen2026skilloptlite}. SkillGrad uses momentum-based
textual gradients, whereas SkillGen induces and tests auditable skills from
successful and failed trajectories
\cite{wang2026skillgrad,ma2026skillgen}. Despite different update rules, they are still
limited to sampled paths.
SkillAdaptor and SkillCAT improve attribution. SkillAdaptor links the first
actionable fault to a targeted revision \cite{yu2026skilladaptor}, whereas
SkillCAT contrasts successful and failed trajectories and validates patches
before merging them into a skill topology \cite{chen2026skillcat}. However,
first-fault analysis yields one correction per rollout, while independent
trajectories may diverge before comparison. \textsc{Branch2Skill} instead
branches from shared prefixes at multiple depths along an elite path. Tree
topology and downstream outcomes isolate each local decision without post hoc
trajectory diagnosis.
\paragraph{Tree search.}
Tree of Thoughts, Reasoning via Planning, and Language Agent Tree Search broaden
test-time reasoning through thought search, Monte Carlo planning, and
environment-guided reflection, respectively
\cite{yao2023tree, zhou2024lats}. Process supervision learns from
intermediate steps: Let's Verify Step by Step and Math-Shepherd train process
reward models, whereas Tree-PLV and SVPO derive tree-based preferences for
verifier training or parameter updates
\cite{lightman2024verify,wang2024mathshepherd,he2024treeplv,
chen2024svpo}. Both lines mainly improve current solutions or model parameters.
Search branches provide finer contrastive evidence. SIGMA uses sibling
comparisons to revise the highest-scoring trajectory
\cite{ren2025sigma}. Contrastive Reasoning Path Synthesis turns high- and
low-quality paths into fine-tuning data \cite{liu2026contrasts}. Branching
Policy Optimization estimates policy advantages from sibling returns
\cite{he2026bpo}. OpenClaw-Skill instead searches candidate skills and uses the
resulting tree for training and inference \cite{lin2026openclawskill}.
\textsc{Branch2Skill} compares shared-prefix task-level continuations across depths and distills outcome differences into a persistent natural-language skill.


\section{Method}
\label{sec:method}

\begin{figure*}[!t]
    \centering
    \includegraphics[width=0.98\textwidth]{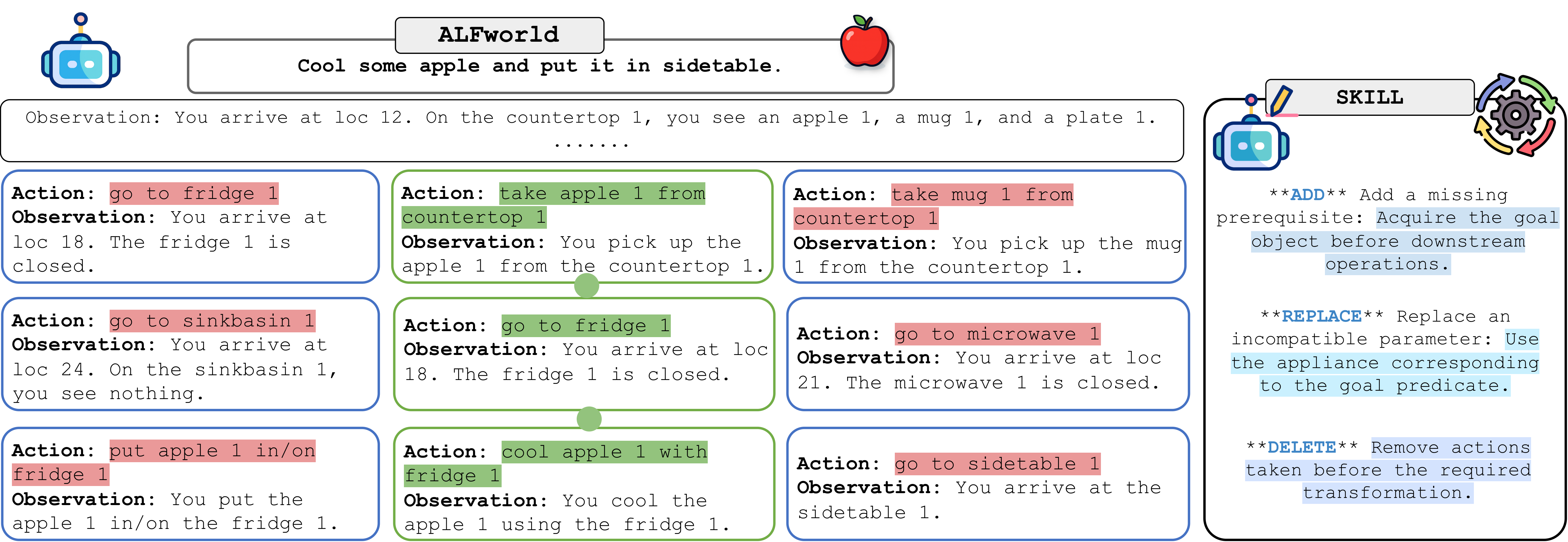}
    \caption{Example of tree-based evidence extraction on an ALFWorld task. Branch2Skill contrasts the correct path with alternative branches that expose distinct execution errors, then converts these local contrasts into Add, Replace, and Delete edits for skill revision.
    }
    \label{fig:alfworld_example}
\end{figure*}

\subsection{Overview}
\label{sec:method_overview}

\textsc{Branch2Skill} updates a persistent natural language skill in discrete
steps. Each update step begins with a small set of $M$ training problems. The
current skill $\mathcal{K}_t$ remains fixed while these problems are processed,
and each problem produces an independent reasoning tree. Evidence from the
resulting trees is then pooled for one skill revision. 

We refer to the model that generates and assesses reasoning nodes as the
\emph{target model}. The model that rewrites the skill is the \emph{skill
model (optimizer)}. As shown in Figure~\ref{fig:method_overview}, the target
model constructs one tree for each problem under a maximum depth
$D_{\max}$, a child count $C$, an iteration budget $I$, and an exploration
coefficient $c_{\mathrm{puct}}$. Once search ends, the method retains an elite
path and the sibling nodes associated with each selected step. The skill model
receives these compact tree slices and produces a candidate skill through
\textsc{Append}, \textsc{Replace}, or \textsc{Delete} operations. A fixed
validation set then determines whether the candidate is accepted or the
previous skill is restored.

\subsection{Stepwise reasoning tree construction}
\label{sec:tree_construction}

For a training problem $x$, the root node $S_0$ represents an empty reasoning
path conditioned on the problem and the current skill $\mathcal{K}_t$. A node
$S_d^i$ denotes the $i$th candidate reasoning step at depth $d$. Thus,
$S_2^1$ denotes the first candidate at the second level, following the
notation used in Figure~\ref{fig:method_overview}. Each nonroot node contains
one reasoning step or one environment action, rather than a complete
solution. When expanding a node, the target model receives the full path from
the root to that node and generates up to $C$ candidate continuations.

Selection uses a policy guided extension of UCT, commonly written as PUCT
\cite{kocsis2006bandit,silver2017mastering}. For a parent node $u$ and one of
its children $v$, the selection score is
\begin{equation}
\operatorname{PUCT}(u,v)
\!=\!
Q(v)
\!+\!
c_{\mathrm{puct}}P(v\!\mid\! u)
\frac{\sqrt{\max(1,N(u))}}{1+N(v)},
\label{eq:puct}
\end{equation}
where $Q(v)$ is the mean score backed up through $v$, $N(\cdot)$ is the visit
count, and $P(v\mid u)$ is the prior assigned when $v$ is created. The
coefficient $c_{\mathrm{puct}}$ controls how strongly the search favours less
visited alternatives. When candidate probabilities are unavailable, children
created during the same expansion receive equal prior mass.

After generating a child, the target model reads the complete reasoning
prefix ending at that child and assigns a scalar progress score. The score
reflects whether the current prefix is likely to support a correct completion.
Model based self assessment has also been used to guide language model tree
search in prior work \cite{yao2023tree}. Each score is then backed up to the root, updating the visit count
and mean score of every node on the selected path.

Search continues until the iteration budget is exhausted, no valid node
remains expandable, or the maximum depth is reached. The target model is
required to complete a path when the remaining depth or search budget becomes
small. This prevents the tree from accumulating intermediate steps without
producing comparable task outcomes. After the main search, valid nonterminal
prefixes may be completed once. These completions do not change the decisions
made during search. They provide terminal outcomes for branches that would
otherwise offer only partial evidence.

\subsection{Shared prefix evidence extraction}
\label{sec:evidence_extraction}

Once a tree has been constructed, \textsc{Branch2Skill} retains one
representative path. A successful terminal path is preferred when the tree
contains one. Otherwise, the method selects the strongest failed path so that
the problem can still provide diagnostic evidence. Terminal outcome is the
primary criterion, while the backed up score, visit count, depth, and
redundancy are used to distinguish paths with the same outcome.

Suppose the retained path contains one selected node at each of $L$ depths.
We denote the path and the sibling set at depth $d$ as
\begin{equation}
\begin{aligned}
\pi^{\star}
&=
\left(
S_1^{i_1^{\star}},
\ldots,
S_L^{i_L^{\star}}
\right), \\
\mathcal{B}_d
&=
\Bigl\{
S_d^j \,\Bigm|\,
\operatorname{pa}(S_d^j)
=
\operatorname{pa}(S_d^{i_d^{\star}}),
j \neq i_d^{\star}
\Bigr\}.
\end{aligned}
\label{eq:sibling_set}
\end{equation}
where $\operatorname{pa}(\cdot)$ denotes the parent of a node. The selected
node $S_d^{i_d^{\star}}$ and every alternative in $\mathcal{B}_d$ share the
same reasoning prefix and first differ at depth $d$. Their downstream outcomes
therefore narrow attribution to the decision made at that depth. This
comparison is more local than a comparison between independently sampled
trajectories, which may differ at several earlier steps.
Each evidence item contains the shared prefix, the selected step, a limited
number of evaluated siblings, their downstream outcomes, and compact search
statistics. The complete tree is not sent to the skill model. No separate
model is used to locate an error after the trajectory has been generated.
Instead, the shared parent and the branch outcomes define the local contrast.

Figure~\ref{fig:alfworld_example} illustrates this process on an ALFWorld task
\cite{shridhar2021alfworld}. The selected route first acquires the target
object and then visits the appliance required by the requested
transformation. Alternatives generated from the same states reveal whether
the existing skill omitted a prerequisite, associated the goal with an
incompatible appliance, or encouraged an action before the required
transformation. Because each comparison begins from the same observation and
action history, the resulting feedback can be tied to a specific decision.


\subsection{Skill revision and validation}
\label{sec:skill_revision}

Evidence from all $M$ trees in the current update step is combined and sent
to the skill model (optimizer) together with the current skill
$\mathcal{K}_t$. The optimizer is called once per update step and returns a
complete candidate skill. It does not produce a collection of uncommitted
suggestions.

The available revisions follow the three operations in
Figure~\ref{fig:method_overview}. \textsc{Append} introduces a missing rule,
condition, or prerequisite. \textsc{Replace} repairs guidance whose action,
object, condition, or scope conflicts with the collected evidence.
\textsc{Delete} removes guidance that is misleading, redundant, or repeatedly
associated with unsuccessful branches. Combining several trees in one update
allows the skill model to distinguish a recurring weakness from an isolated
mistake.

The resulting candidate $\widehat{\mathcal{K}}_{t+1}$ is evaluated on a fixed
validation set $\mathcal{V}$ using direct inference without tree search. The
current skill and the candidate skill are evaluated under the same conditions.
Let $\mathbf{g}(\mathcal{K};\mathcal{V})$ denote the vector of designated
validation metrics. The validation gate applies the following componentwise
nonregression rule:
\begin{equation}
\mathcal{K}_{t+1}
=
\begin{cases}
\widehat{\mathcal{K}}_{t+1},
&
\mathbf{g}(\widehat{\mathcal{K}}_{t+1};\mathcal{V})
\succeq
\mathbf{g}(\mathcal{K}_t;\mathcal{V}),
\\[4pt]
\mathcal{K}_t,
&
\text{otherwise},
\end{cases}
\label{eq:validation_gate}
\end{equation}
where $\succeq$ requires that none of the designated validation metrics
decrease. A candidate that fails this condition is discarded, and the
previous skill remains active. A candidate that preserves or improves
validation performance becomes the fixed skill for the next update step. The
method also retains the checkpoint with the strongest validation performance
for final evaluation. This gate prevents a locally plausible revision from
persisting when it reduces the broader utility of the skill.


\section{Experiments}
\label{sec:experiments}

\subsection{Experimental setup}
\label{sec:experimental_setup}

\paragraph{Benchmarks.}
We evaluate \textsc{Branch2Skill} on six benchmarks. SearchQA tests
noisy-evidence question answering \cite{dunn2017searchqa}, SpreadsheetBench
tests realistic workbook manipulation \cite{ma2024spreadsheetbench}, and
OfficeQA tests enterprise-document reasoning using the SkillOpt-aligned split
\cite{opsahlong2026officeqa,yang2026skillopt}. DocVQA covers document-image
question answering \cite{mathew2021docvqa}, LiveMathematicianBench (LiveMath)
covers research-level mathematics \cite{he2026livemath}, and ALFWorld covers
embodied planning and tool use \cite{shridhar2021alfworld}.
Most baseline results are taken from SkillOpt. We reran SkillOpt using its official implementation and released benchmark-specific training configurations, and report the resulting performance and evolution-token costs.
Branch2Skill run performs a single evolution epoch. Table~\ref{tab:benchmark_protocol} reports the splits and evolution
costs.

\begin{table}[t]
    \centering

    \small
    \setlength{\tabcolsep}{3.5pt}
    \begin{tabular}{lrrrr}
        \toprule
        Benchmark & Train & Val. & Test & Tokens (M) \\
        \midrule
        SearchQA          & 40 & 200 & 1400 & 27.5 \\
        SpreadsheetBench  & 40 &  40 &  280 & 17.7 \\
        OfficeQA          & 20 &  24 &  172 & 13.4 \\
        DocVQA            & 80 &  53 &  374 & 29.3 \\
        LiveMath          & 20 &  40 &  124 &  6.7 \\
        ALFWorld          & 20 &  18 &  134 & 46.7 \\
        \midrule
        Total             &    &     &      & 141.3 \\
        \bottomrule
    \end{tabular}
    \caption{Dataset split and training consumption}
    \label{tab:benchmark_protocol}
\end{table}


\newcommand{\gain}[1]{\,+#1}
\newcommand{\loss}[1]{\,-#1}
\newcommand{\same}{\,\(\pm\)0.0}
\newcommand{\best}[1]{\textbf{#1}}
\newcommand{\second}[1]{\underline{#1}}

\begin{table*}[!t]
\centering
\begingroup
\small
\setlength{\tabcolsep}{2.4pt}
\renewcommand{\arraystretch}{1.08}
\begin{tabular*}{\textwidth}{@{\extracolsep{\fill}}llrrrrrr@{}}
\toprule
\textbf{Model} & \textbf{Skill Source}
& \multicolumn{1}{c}{\textbf{SearchQA}}
& \multicolumn{1}{c}{\textbf{Spreadsheet}}
& \multicolumn{1}{c}{\textbf{OfficeQA}}
& \multicolumn{1}{c}{\textbf{DocVQA}}
& \multicolumn{1}{c}{\textbf{LiveMath}}
& \multicolumn{1}{c}{\textbf{ALFWorld}} \\
\midrule
\multirow{8}{*}{GPT{-}5.5} & \textit{No skill} & 77.7 & 41.8 & 33.1 & 78.8 & 37.6 & 83.6 \\
\cmidrule(lr){2-8}
  & Human skill & 81.8\gain{4.1} & 72.9\gain{31.1} & 66.9\gain{33.8} & 90.1\gain{11.3} & 38.4\gain{0.8} & 91.8\gain{8.2} \\
  & LLM skill & 80.9\gain{3.2} & 43.2\gain{1.4} & 51.7\gain{18.6} & 89.6\gain{10.8} & 40.0\gain{2.4} & 93.3\gain{9.7} \\
  & Trace2Skill & 82.4\gain{4.7} & 49.6\gain{7.8} & 65.7\gain{32.6} & 90.6\gain{11.8} & 52.0\gain{14.4} & 87.3\gain{3.7} \\
  & TextGrad & 81.4\gain{3.7} & 41.1\loss{0.7} & 42.0\gain{8.9} & 87.2\gain{8.4} & 49.2\gain{11.6} & 82.8\loss{0.8} \\
  & GEPA & \best{84.8}\gain{7.1} & 73.6\gain{31.8} & 63.9\gain{30.8} & 89.1\gain{10.3} & 43.2\gain{5.6} & 85.8\gain{2.2} \\
  & SkillOpt & 80.4\gain{2.7} & \second{76.4}\gain{34.6} & \best{72.1}\gain{39.0} & \second{91.2}\gain{12.4} & \second{54.8}\gain{17.2} & \second{95.5}\gain{11.9} \\
  & \textbf{Branch2Skill (Ours)} & \second{83.9}\gain{6.2} & \best{77.5}\gain{35.7} & \second{70.9}\gain{37.8} & \best{92.2}\gain{13.4} & \best{60.5}\gain{22.9} & \best{96.3}\gain{12.7} \\
\midrule
\multirow{8}{*}{GPT{-}5.4} & \textit{No skill} & 76.9 & 41.4 & 50.0 & 77.6 & 36.8 & 75.4 \\
\cmidrule(lr){2-8}
  & Human skill & 80.1\gain{3.2} & 59.3\gain{17.9} & 59.3\gain{9.3} & 88.5\gain{10.9} & 41.6\gain{4.8} & 74.6\loss{0.8} \\
  & LLM skill & 79.4\gain{2.5} & 46.1\gain{4.7} & 20.4\loss{29.6} & \second{90.4}\gain{12.8} & 36.8\same & 83.6\gain{8.2} \\
  & Trace2Skill & 81.2\gain{4.3} & 53.2\gain{11.8} & 51.2\gain{1.2} & 89.3\gain{11.7} & 40.0\gain{3.2} & 73.1\loss{2.3} \\
  & TextGrad & \second{82.5}\gain{5.6} & 38.6\loss{2.8} & 54.7\gain{4.7} & 85.3\gain{7.7} & 36.6\loss{0.2} & 78.4\gain{3.0} \\
  & GEPA & 82.4\gain{5.5} & 61.1\gain{19.7} & 60.4\gain{10.4} & 88.3\gain{10.7} & 41.6\gain{4.8} & 74.6\loss{0.8} \\
  & SkillOpt & 78.5\gain{1.6} & \second{61.4}\gain{20.0} & \second{62.8}\gain{12.8} & \second{90.4}\gain{12.8} & \second{44.0}\gain{7.2} & \second{91.0}\gain{15.6} \\
  & \textbf{Branch2Skill (Ours)} & \best{82.8}\gain{5.9} & \best{66.8}\gain{25.4} & \best{65.1}\gain{15.1} & \best{91.2}\gain{13.6} & \best{50.8}\gain{14.0} & \best{91.8}\gain{16.4} \\
\midrule
\multirow{8}{*}{GPT{-}5.4{-}mini} & \textit{No skill} & 75.9 & 36.1 & 22.1 & 71.4 & 14.7 & 73.1 \\
\cmidrule(lr){2-8}
  & Human skill & 77.2\gain{1.3} & 42.9\gain{6.8} & 45.9\gain{23.8} & 85.0\gain{13.6} & 28.8\gain{14.1} & 56.7\loss{16.4} \\
  & LLM skill & 76.1\gain{0.2} & 36.8\gain{0.7} & 36.6\gain{14.5} & 86.4\gain{15.0} & 28.0\gain{13.3} & 65.7\loss{7.4} \\
  & Trace2Skill & 78.6\gain{2.7} & 40.7\gain{4.6} & 20.9\loss{1.2} & 88.5\gain{17.1} & \second{32.8}\gain{18.1} & 82.8\gain{9.7} \\
  & TextGrad & 77.5\gain{1.6} & 38.2\gain{2.1} & 30.0\gain{7.9} & 84.0\gain{12.6} & 27.2\gain{12.5} & 70.9\loss{2.2} \\
  & GEPA & 79.4\gain{3.5} & 42.5\gain{6.4} & 45.3\gain{23.2} & 83.7\gain{12.3} & 27.2\gain{12.5} & 81.3\gain{8.2} \\
  & SkillOpt & \best{80.2}\gain{4.3} & \second{47.5}\gain{11.4} & \second{48.8}\gain{26.7} & \second{90.9}\gain{19.5} & \second{32.8}\gain{18.1} & \second{85.8}\gain{12.7} \\
  & \textbf{Branch2Skill (Ours)} & \second{79.5}\gain{3.6} & \best{49.3}\gain{13.2} & \best{50.6}\gain{28.5} & \best{91.2}\gain{19.8} & \best{40.3}\gain{25.6} & \best{87.3}\gain{14.2} \\
\midrule
\multirow{8}{*}{GPT{-}5.4{-}nano} & \textit{No skill} & 55.8 & 23.5 & 16.3 & 30.8 & 23.2 & 34.3 \\
\cmidrule(lr){2-8}
  & Human skill & 66.1\gain{10.3} & 41.8\gain{18.3} & \best{46.5}\gain{30.2} & 73.5\gain{42.7} & 16.8\loss{6.4} & 29.9\loss{4.4} \\
  & LLM skill & 62.4\gain{6.6} & 33.6\gain{10.1} & 12.8\loss{3.5} & 71.7\gain{40.9} & 20.0\loss{3.2} & 65.7\gain{31.4} \\
  & Trace2Skill & 69.9\gain{14.1} & 35.4\gain{11.9} & 16.3\same & 77.3\gain{46.5} & 25.6\gain{2.4} & 58.2\gain{23.9} \\
  & TextGrad & 73.4\gain{17.6} & 32.5\gain{9.0} & 38.7\gain{22.4} & 68.3\gain{37.5} & 20.8\loss{2.4} & 42.5\gain{8.2} \\
  & GEPA & 73.2\gain{17.4} & 37.2\gain{13.7} & \second{45.0}\gain{28.7} & 71.2\gain{40.4} & 24.8\gain{1.6} & 68.7\gain{34.4} \\
  & SkillOpt & \second{74.8}\gain{19.0} & \second{42.5}\gain{19.0} & 12.2\loss{4.1} & \second{80.2}\gain{49.4} & \second{27.2}\gain{4.0} & \second{69.4}\gain{35.1} \\
  & \textbf{Branch2Skill (Ours)} & \best{80.1}\gain{24.3} & \best{48.2}\gain{24.7} & 29.7\gain{13.4} & \best{84.2}\gain{53.4} & \best{31.5}\gain{8.3} & \best{71.6}\gain{37.3} \\
\midrule
\multirow{8}{*}{\shortstack[l]{Qwen3.6{-}\\35B{-}A3B}} & \textit{No skill} & 72.7 & 38.2 & 45.9 & 87.6 & 31.2 & 59.7 \\
\cmidrule(lr){2-8}
  & Human skill & 74.1\gain{1.4} & 44.3\gain{6.1} & 41.9\loss{4.0} & 87.4\loss{0.2} & 29.6\loss{1.6} & 44.8\loss{14.9} \\
  & LLM skill & 72.6\loss{0.1} & 42.9\gain{4.7} & 46.5\gain{0.6} & 88.4\gain{0.8} & 24.8\loss{6.4} & 60.4\gain{0.7} \\
  & Trace2Skill & 75.4\gain{2.7} & 33.2\loss{5.0} & 32.0\loss{13.9} & 90.4\gain{2.8} & 29.6\loss{1.6} & 70.9\gain{11.2} \\
  & TextGrad & 76.4\gain{3.7} & 22.9\loss{15.3} & 33.7\loss{12.2} & 84.5\loss{3.1} & 7.2\loss{24.0} & 67.9\gain{8.2} \\
  & GEPA & 75.8\gain{3.1} & 45.4\gain{7.2} & 43.6\loss{2.3} & 88.0\gain{0.4} & 31.2\same & \second{73.9}\gain{14.2} \\
  & SkillOpt & \second{80.3}\gain{7.6} & \second{47.5}\gain{9.3} & \second{47.1}\gain{1.2} & \second{91.4}\gain{3.8} & \second{41.6}\gain{10.4} & \best{82.1}\gain{22.4} \\
  & \textbf{Branch2Skill (Ours)} & \best{80.4}\gain{7.7} & \best{49.6}\gain{11.4} & \best{50.0}\gain{4.1} & \best{92.5}\gain{4.9} & \best{45.2}\gain{14.0} & \best{82.1}\gain{22.4} \\
\bottomrule
\end{tabular*}
\endgroup
\caption{\textbf{Performance across six benchmarks and five target
    models.} Values following each score report the absolute percentage point
    change relative to \emph{No skill} for the same target model. Bold and
    underlined entries denote the best and second-best results within each
    model group. All metrics are higher is better. GPT-5.5 is the skill model
    for every \textsc{Branch2Skill} result in this table.}
\label{tab:main_results}
\end{table*}

\begin{figure*}[!t]
    \centering
    \begin{subfigure}[t]{0.48\textwidth}
        \centering
        \includegraphics[width=0.66\linewidth]{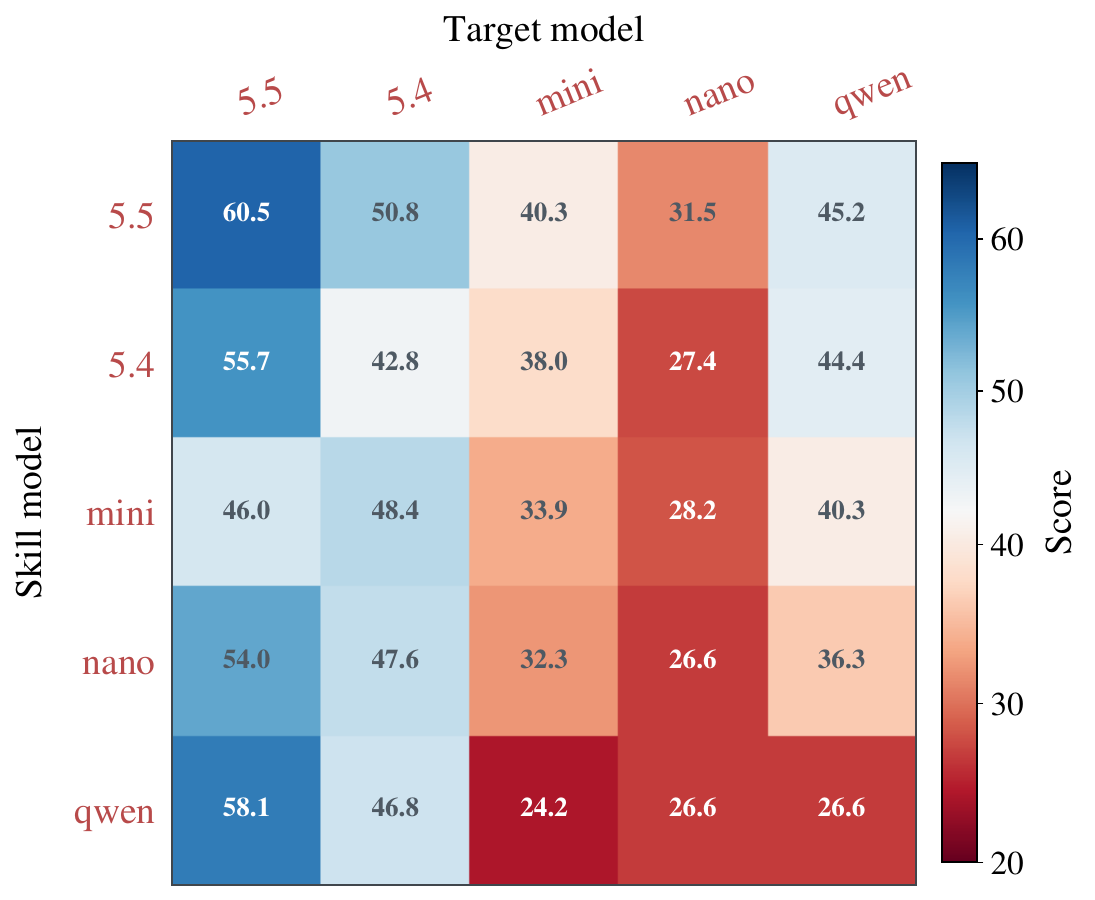}
        \caption{LiveMath}
        \label{fig:model_role_livemath}
    \end{subfigure}
    \hfill
    \begin{subfigure}[t]{0.48\textwidth}
        \centering
        \includegraphics[width=0.66\linewidth]{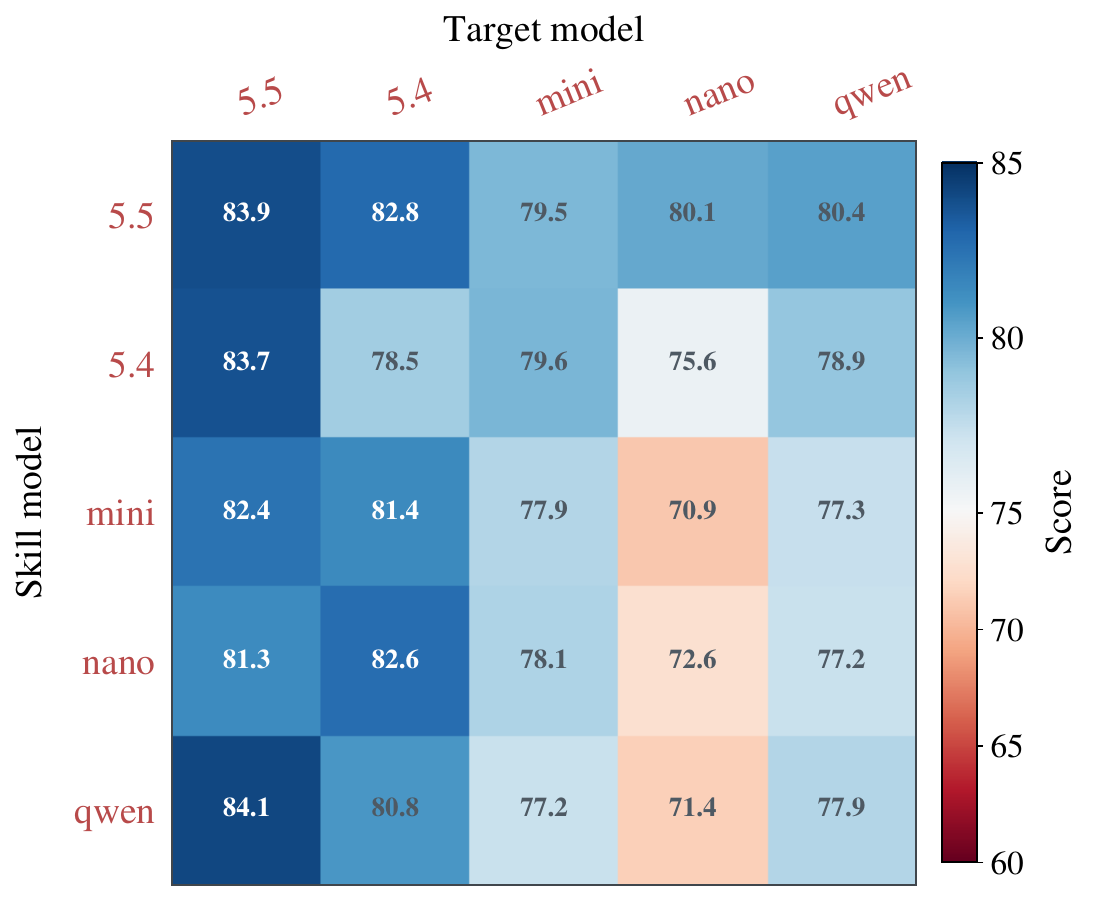}
        \caption{SearchQA}
        \label{fig:model_role_searchqa}
    \end{subfigure}
    \caption{\textbf{Interaction between the target model and the skill model.}}
    \label{fig:model_role_matrix}
\end{figure*}

\paragraph{Models and evolution protocol.}
Except in the model-role analysis, GPT-5.5 serves as the skill model. Target
models are GPT-5.5, GPT-5.4, GPT-5.4-mini, GPT-5.4-nano, and
Qwen3.6-35B-A3B. OpenAI models use medium reasoning effort, whereas Qwen uses
its standard configuration.

For each target--benchmark pair, every update uses two training problems and
builds an independent reasoning tree for each. Defaults are three children per
expansion, depth six, 40 iterations per problem. The target model generates candidate steps and
estimates prefix success. After search, valid nonterminal prefixes are
completed to provide terminal evidence.

The skill model receives the elite path and same-parent siblings at retained
depths, then proposes one candidate using \textsc{Append}, \textsc{Replace},
and \textsc{Delete} (Section~\ref{sec:skill_revision}). The candidate is
evaluated directly on the validation set without tree search. Lower scores
trigger rollback; nondecreasing scores are accepted. Final evaluation uses the
best-validation checkpoint.

\paragraph{Baselines and metrics.}
We compare against the target model without a skill, a human-written skill,
and a skill generated once from the model's parametric knowledge. We also
include Trace2Skill \cite{ni2026trace2skill}, TextGrad
\cite{yuksekgonul2025textgrad}, GEPA \cite{agrawal2026gepa}, and SkillOpt
\cite{yang2026skillopt}. We follow the task interfaces and scoring protocol used by SkillOpt.

\subsection{Main results}
\label{sec:main_results}
\textsc{Branch2Skill} led Table~\ref{tab:main_results}, ranking best or tied in
26 of 30 settings and outperforming SkillOpt in 27. It raised the mean score
from 50.8 to 69.1, compared with 66.0 for SkillOpt, yielding average gains of
18.3 and 15.3 points. Its advantage was positive on every benchmark, peaking
at 5.6 points on LiveMath and 4.7 on OfficeQA. With GPT-5.4-nano, it exceeded
no skill by 26.9 points and SkillOpt by 6.5 points. The few deficits were at
most 1.2 points. Although the human-written skill remained best on OfficeQA
with GPT-5.4-nano, \textsc{Branch2Skill} still surpassed no skill by 13.4
points and SkillOpt by 17.5 points.

\paragraph{Evolution cost.}
Across six GPT-5.5 experiments, \textsc{Branch2Skill} used 141.3 million
evolution tokens, 73.2\% fewer than SkillOpt, while increasing the aggregate
test gain from 117.8 to 128.7 points. Gain per million tokens rose from 0.224
to 0.911, a 4.07-fold increase. On LiveMath, token use fell from 23.2 million
to 6.7 million, a 71.1\% reduction. These counts include all evolution model
calls (Figure~\ref{fig:efficiency}).

\FloatBarrier

\subsection{Interaction between target and skill models}
\label{sec:model_role_analysis}

The skill and target models need not match. Across all 25 pairings
(Figure~\ref{fig:model_role_matrix}), GPT-5.5 was the strongest skill model,
averaging 45.7 on LiveMath versus 41.7 for the runner-up, and 81.3 versus 79.3
on SearchQA. It produced the best LiveMath skill for every target, by
2.8--9.3 points. SearchQA was less sensitive: Qwen with a GPT-5.5 target
scored 84.1 versus 83.9 diagonally, while GPT-5.4 with GPT-5.4-mini scored
79.6 versus 79.5. Thus, branch evidence transfers across model identities,
although difficult mathematical reasoning benefits more from a stronger skill
model.

\subsection{Ablation study}
\label{sec:ablation}

%








\begin{table}[H]
\centering
\begingroup
\small
\setlength{\tabcolsep}{2.6pt}
\renewcommand{\arraystretch}{1.06}
\begin{tabular*}{\columnwidth}{@{\extracolsep{\fill}}llrr@{}}
\toprule
\textbf{Factor} & \textbf{Setting}
& \multicolumn{1}{c}{\textbf{SearchQA}}
& \multicolumn{1}{c}{\textbf{LiveMath}} \\
\midrule
\multirow{3}{*}{Maximum depth}
  & 2 & 78.5 & 52.4 \\
  & 4 & 80.3 & 57.3 \\
  & 6 & \textbf{83.9} & \textbf{60.5} \\
\midrule
\multirow{3}{*}{Child-node count}
  & 1 & 79.3 & 53.2 \\
  & 2 & 81.3 & 56.5 \\
  & 3 & \textbf{83.9} & \textbf{60.5} \\
\midrule
\multirow{3}{*}{Iteration budget}
  & 10 & 78.3 & 57.3 \\
  & 20 & 81.6 & 58.9 \\
  & 40 & \textbf{83.9} & \textbf{60.5} \\
\midrule
\multirow{3}{*}{Selection method}
  & Single path & 81.0 & 46.8 \\
  & Whole tree & 78.3 & 45.2 \\
  & \textbf{Branch2Skill (Ours)} & \textbf{83.9} & \textbf{60.5} \\
\bottomrule
\end{tabular*}
\endgroup
\caption{\textbf{Ablation of search capacity and evidence selection.}
    Each block varies one factor while holding the others at their defaults:
    maximum depth 6, three child nodes, an iteration budget of 40, and our
    evidence selection method. Both the target and skill models use GPT-5.5.
    \emph{Single path} excludes sibling evidence; \emph{Whole tree} sends all
    nodes to the optimizer.}
    \label{tab:search_ablation}
\end{table}

Table~\ref{tab:search_ablation} shows that broader search improved both
benchmarks. On SearchQA/LiveMath, depth two-to-six raised scores from
78.5/52.4 to 83.9/60.5, children one-to-three added 4.6/7.3 points, and
iterations 10-to-40 added 5.6/3.2 points. Evidence selection mattered more:
ours beat the elite path/whole tree by 13.7/15.3 points on LiveMath and
2.9/5.6 on SearchQA. Whole-tree underperformance on both benchmarks supports
same-parent comparisons over unfiltered nodes.
\FloatBarrier

\begin{figure*}[!t]
  \centering
  \includegraphics[width=0.94\textwidth]{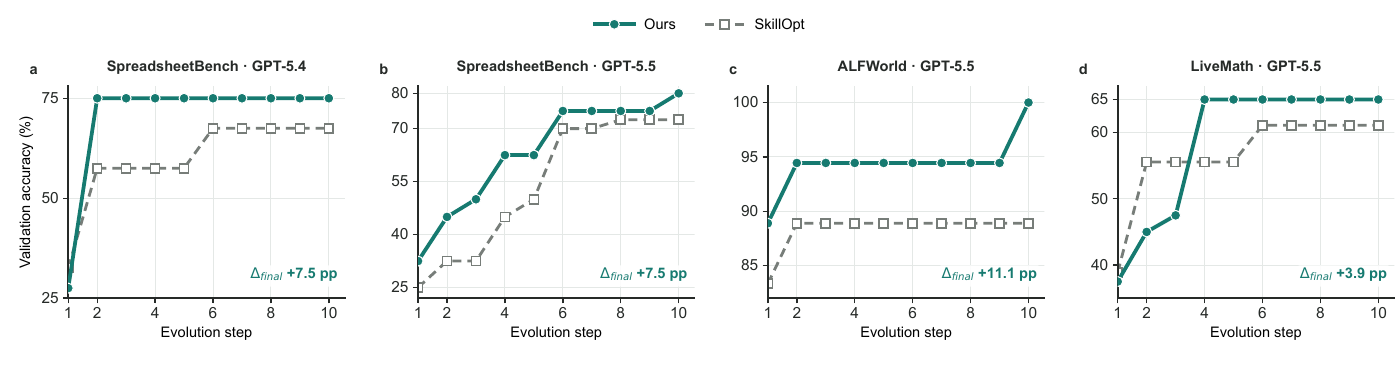}
  \caption{\textbf{Validation performance during skill evolution.} token
  efficiency is reported separately in Figure~\ref{fig:efficiency}.}
  \label{fig:evolution_dynamics}
\end{figure*}

\subsection{Transfer across models and datasets}
\label{sec:transfer}


\begin{table}[H]
\centering
\begingroup
\small
\setlength{\tabcolsep}{2.2pt}
\renewcommand{\arraystretch}{1.08}
\begin{tabular*}{\columnwidth}{@{\extracolsep{\fill}}llrrr@{}}
\toprule
\multicolumn{5}{@{}l}{\textbf{(a) Cross-model transfer from GPT-5.4}} \\
\addlinespace[2pt]
\textbf{Benchmark} & \textbf{Target model}
& \multicolumn{1}{c}{\textbf{No skill}}
& \multicolumn{1}{c}{\textbf{With skill}}
& \multicolumn{1}{c}{\(\Delta\)} \\
\midrule
\multirow{2}{*}{LiveMath}
  & GPT-5.5 & 37.6 & \textbf{56.5} & +18.9 \\
  & GPT-5.4-mini & 14.7 & \textbf{36.3} & +21.6 \\
\addlinespace[1pt]
\multirow{2}{*}{Spreadsheet}
  & GPT-5.5 & 41.8 & \textbf{73.6} & +31.8 \\
  & GPT-5.4-mini & 36.1 & \textbf{46.8} & +10.7 \\
\midrule
\multicolumn{5}{@{}l}{\textbf{(b) Cross-dataset transfer}} \\
\addlinespace[2pt]
\textbf{Target model} & \textbf{Target dataset}
& \multicolumn{1}{c}{\textbf{No skill}}
& \multicolumn{1}{c}{\textbf{With skill}}
& \multicolumn{1}{c}{\(\Delta\)} \\
\midrule
GPT-5.4 & \multirow{3}{*}{Omni-MATH} & 79.8 & \textbf{83.3} & +3.5 \\
GPT-5.4-mini & & 70.4 & \textbf{73.2} & +2.8 \\
GPT-5.4-nano & & 69.8 & \textbf{72.6} & +2.8 \\
\bottomrule
\end{tabular*}
\endgroup
\caption{\textbf{Cross-model and cross-dataset transfer.}
    In panel (a), a skill evolved with GPT-5.4 as the target model and
    GPT-5.5 as the skill model is frozen and applied to other target models.
    In panel (b), a skill evolved from 40 OlympiadBench examples is frozen and
    evaluated on Omni-MATH. No target-domain update is performed after
    transfer. $\Delta$ is the absolute percentage point gain over
    \emph{No skill}.}
\label{tab:transfer}
\end{table}


The frozen skills transfer across target models without further evolution. A
LiveMath skill evolved from GPT-5.4 trajectories improves GPT-5.5 and
GPT-5.4-mini by 18.9 and 21.6 points, respectively. The corresponding
SpreadsheetBench gains are 31.8 and 10.7 points.
A mathematical skill evolved from 40 OlympiadBench problems
\cite{he2024olympiadbench} also transfers to Omni-MATH
\cite{gao2025omnimath} without further optimization. It improves GPT-5.4,
GPT-5.4-mini, and GPT-5.4-nano by 3.5, 2.8, and 2.8 points, respectively.
Although smaller than in-domain gains, these improvements are consistent with
learning reusable procedures rather than memorizing example formats.

\subsection{Evolution dynamics}
\label{sec:evolution_dynamics}

The validation gate yields nondecreasing retained-skill curves in
Figure~\ref{fig:evolution_dynamics}. \textsc{Branch2Skill} finishes 7.5 points
above SkillOpt on SpreadsheetBench with both GPT-5.4 and GPT-5.5, 11.1 points
ahead on ALFWorld, and 3.9 on LiveMath with GPT-5.5. Several curves plateau
because the gate rejects later revisions unless validation performance is
preserved. Thus, the curves track retained skills, whereas the token comparison
measures their evolution cost.

\section{Conclusion}
\label{sec:conclusion}


\textsc{Branch2Skill} turns reasoning-tree search into supervision for
persistent skill evolution by contrasting an elite path with shared-prefix
siblings and retaining only updates that preserve validation performance.
Across six benchmarks and five target models, it improves every no-skill
baseline; with GPT-5.5, it uses $73.2\%$ fewer evolution tokens than SkillOpt
while achieving a $9.3\%$ larger aggregate gain. Ablations show that structured
sibling evidence outperforms single-path and whole-tree variants, indicating
that temporary search trees can capture local decisions as reusable skills for
future reasoning.

\appendix




\FloatBarrier
\balance
{\fontsize{7.5}{8.4}\selectfont
\bibliographystyle{ieeenat_fullname}
\bibliography{aaai2027}
}


\end{document}